\documentclass[runningheads]{llncs}
 
\usepackage{eccv}

\usepackage{eccvabbrv}

\usepackage{graphicx}
\usepackage{booktabs}

\usepackage{adjustbox}
\usepackage{array}
\usepackage{multirow}
\usepackage{pifont}

\usepackage[accsupp]{axessibility}  

\usepackage{caption}
\usepackage{subcaption}

\usepackage{hyperref}

\usepackage{orcidlink}

\usepackage{amsmath}
\DeclareMathOperator*{\argmax}{arg\,max}
\DeclareMathOperator*{\argmin}{arg\,min}
\usepackage{bbm}

\begin{document}

\newcommand{\additionalInfo}[1]{#1}
\newcommand{\RPN}{\mathrm{RPN}}
\newcommand{\Refi}{\mathrm{Ref}}
\newcommand{\pan}{\mathrm{pan}}
\newcommand{\sem}{\mathrm{sem}}
\newcommand{\mask}{\mathrm{msk}}
\newcommand{\bbox}{\mathrm{box}}
\newcommand{\bboxc}{\mathrm{cls}}
\newcommand{\cntr}{\mathrm{cntr}}
\newcommand{\offs}{\mathrm{offs}}

\newcommand{\todo}[1]{{\textcolor{red}{[#1]}}}
\newcommand{\highl}[1]{{\textcolor{blue}{#1}}}

\title{Language-Guided Instance-Aware Domain-Adaptive Panoptic Segmentation} 

\titlerunning{Abbreviated paper title}

\author{Elham Amin Mansour\inst{1}, 
Ozan Unal\inst{1},
Suman Saha\inst{2}, 
Benjamin Bejar\inst{2}, \and \\ 
Luc Van Gool\inst{1} }
\authorrunning{Elham A. et al.}
\titlerunning{LIDAPS}


\institute{Computer Vision Lab, ETH Zurich \and
Swiss Data Science Center, PSI \\
\email{\{elham.aminmansour\}@inf.ethz.ch}}

\maketitle

\begin{abstract}

The increasing relevance of panoptic segmentation is tied to the advancements in autonomous driving and AR/VR applications. However, the deployment of such models has been limited due to the expensive nature of dense data annotation, giving rise to unsupervised domain adaptation (UDA). A key challenge in panoptic UDA is reducing the domain gap between a labeled source and an unlabeled target domain while harmonizing the subtasks of semantic and instance segmentation to limit catastrophic interference. While considerable progress has been achieved, existing approaches mainly focus on the adaptation of semantic segmentation. In this work, we focus on incorporating instance-level adaptation via a novel instance-aware cross-domain mixing strategy IMix. IMix significantly enhances the panoptic quality by improving instance segmentation performance. Specifically, we propose inserting high-confidence predicted instances from the target domain onto source images, retaining the exhaustiveness of the resulting pseudo-labels while reducing the injected confirmation bias. Nevertheless, such an enhancement comes at the cost of degraded semantic performance, attributed to catastrophic forgetting. To mitigate this issue, we regularize our semantic branch by employing CLIP-based domain alignment (CDA), exploiting the domain-robustness of natural language prompts. Finally, we present an end-to-end model incorporating these two mechanisms called LIDAPS, achieving state-of-the-art results on all popular panoptic UDA benchmarks. 

\end{abstract}
\vspace{-12px} \section{Introduction}\label{sec:intro}

Panoptic segmentation~\cite{Kirillov_2019_CVPR} unifies semantic and instance segmentation by not only assigning a class label to each pixel but also segmenting each object into its own instance. The common approach when tackling panoptic segmentation is to deconstruct it into two subtasks and later fuse the resulting dense predictions~\cite{cvrn, edaps}. The challenge in such an approach lies in the contradictory nature of the individual task objectives~\cite{cvrn}.
While semantic segmentation seeks to map the embeddings of semantically similar object instances into a class-specific representation, instance segmentation aims to learn discriminative features to separate instances from one another, resulting in conflicting gradients from two different objectives. Despite the apparent challenges, the rich semantic information with instance-level discrimination is crucial for downstream applications such as autonomous driving or AR/VR. Yet, the complexity and cost of acquiring such panoptic annotations heavily hinder the real-world deployability of such models. Furthermore, given the variance in data distribution between different domains caused by geographical changes, object selection, weather conditions, or sensor setups, models trained on previously acquired annotated data often perform poorly in new domains. This phenomenon, known as the ``domain gap'', remains a further limiting factor. To this end, recent works have focused on incorporating data-efficiency into panoptic segmentation through the task of unsupervised domain adaptation (UDA)~\cite{cvrn, unidaformer, edaps}. In contrast to the aforementioned supervised setting, in panoptic UDA a model is trained on labeled source domain images and unlabeled target domain images with supervision only available on the source domain. This allows (i) available labeled data to be used to tackle further domains (real-to-real adaptation) or (ii) to reduce annotation requirements altogether (synthetic-to-real adaptation).

\begin{figure*}[t]
  \centering
\includegraphics[width=\textwidth]{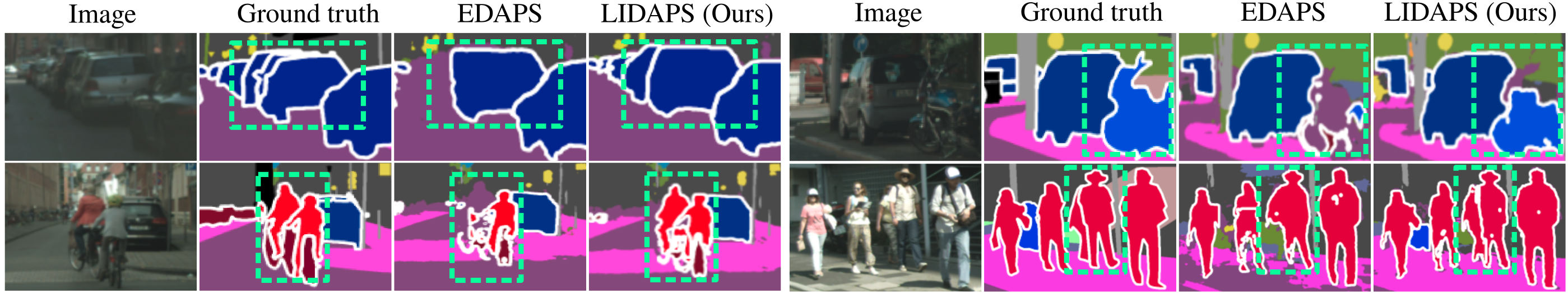}
  \caption{ While previous SOTA methods for panoptic UDA such as EDAPS~\cite{edaps} achieve good semantic segmentation performance, they struggle to predict correct object boundaries and thus instance segmentation masks.
  }
  \label{fig:intro_teaser}
\vspace{-0px}\end{figure*}

\begin{figure}[t]
    \centering
    \captionof{table}{Comparison of LIDAPS with SOTA on different aspects such as self-training (ST) type; ST feature space: semantic (Sem) vs. instance (Inst);  shared (SR) vs. task-specific (TR) representations;
    sampling strategies: ClassMix \cite{classmix} vs. proposed IMix (Sec. \ref{sec:imix}); and proposed CLIP-based domain alignment (CDA).}
    \vspace{4mm} 
    \label{table:intro-comp-sota-aspects}
    \resizebox{0.93\textwidth}{!}{%
    \begin{tabular}{l|cccccccc}
    \toprule
    Method & ST & SemST & InstST & SR & TR & ClassMix & IMix & CDA \\
    \midrule
    CVRN \cite{cvrn} & Offline & \checkmark & \checkmark & & \checkmark & & & \\
    UniDAformer \cite{unidaformer} & Online & \checkmark & \checkmark & \checkmark & & & & \\
    EDAPS \cite{edaps} & Online & \checkmark & & \checkmark & \checkmark & \checkmark & & \\
    \textbf{LIDAPS (Ours)} & Online & \checkmark & \checkmark & \checkmark & \checkmark & \checkmark & \checkmark & \checkmark \\
    \bottomrule
    \end{tabular}
    }
\vspace{-5px} \end{figure}

However, under a panoptic UDA setting, balancing both tasks and limiting the effects that arise from the contradictory objectives becomes more challenging due to the lack of a supervisory signal on the target domain.
{In Tab.~\ref{table:intro-comp-sota-aspects}, we provide an overview of SOTA panoptic UDA methods \cite{cvrn,unidaformer,edaps} based on different criteria.}
Apart from CVRN~\cite{cvrn} that avoids the problem by completely decoupling the two tasks and training individual networks, {i.e., fully rely on task-specific representations (TR)}, previously proposed methods that utilize more memory-efficient unified network architecture (e.g. exploiting both shared (SR) and task-specific representations) have tackled panoptic UDA by only adapting the semantic segmentation branch to improve panoptic quality. Specifically, EDAPS~\cite{edaps} utilizes ClassMix~\cite{classmix} to generate semantically cross-domain mixed inputs that align the target domain to the source, and, UniDAformer~\cite{unidaformer} hierarchically calibrates the semantic masks across generated regions, superpixels, and pixels. Such SOTA methods for panoptic UDA are thus able to learn good semantic segmentation masks in the target domain, however, are prone to predict inaccurate instance segmentation masks due to the conflicting objectives. This problem is more prominent when multiple overlapping or occluded object instances are present in a scene. An example is shown in Fig.~\ref{fig:intro_teaser} where it can be seen that while EDAPS correctly predicts the semantic segmentation masks for the ``car'' (top-left) and ``person'' (bottom-right) classes,
it fails to identify individual instance boundaries resulting in the merging of objects. 
This limitation is expected, given the lack of adaptation for instance segmentation. In the current literature, the adaptation on an instance-level for panoptic UDA is heavily underexplored~\cite{cvrn, unidaformer}, with no work in instance-level cross-domain mixing.

In this work, we propose a novel instance-aware mixing strategy IMix (Sec. \ref{sec:imix}), to improve the recognition quality of a panoptic UDA model directly. With IMix, we leverage the panoptic predictions of a model to generate a cross-domain input image consisting of high-confidence instances from the target domain pasted onto a source image and finetune itself through self-supervision. By employing target-to-source mixing, we retain the exhaustiveness of the generated panoptic pseudo-label, i.e. each object within the scene always has an associated instance label. This allows us to reduce the \textit{confirmation bias} while directly learning target instance segmentation on a simpler source background.

While IMix enhances panoptic quality via improved instance segmentation performance, the enhancement is limited due to a drop in semantic segmentation performance (Table \ref{TableAblateAll}). The model finetuned with IMix becomes subject to catastrophic interference, yielding the ability to map semantically similar objects into a joint embedding in favor of increased instance separability~\cite{daformer}. To remedy this, we propose employing CLIP-based domain alignment (CDA) to act as a regularizer on the semantic branch (Sec. \ref{sec:cda}). In essence, CDA continually aligns both the target and source domains with a pre-trained frozen CLIP~\cite{clip} model. Specifically, we leverage the rich feature space of CLIP to construct class-wise embeddings from a set of static text prompts. We then compute their inner product with the semantic decoder features to generate per-pixel-text similarity maps following DenseCLIP~\cite{DenseClip} that are then supervised via ground truth or pseudo-target labels. 

Finally, we combine our two proposed modules with a unified transformer backbone and individual task decoders to construct LIDAPS, a language-guided instance-aware domain-adapted panoptic segmentation model. {Our proposed LIDAPS, while improving instance segmentation, is also able to enhance the semantic quality through CDA. For example, LIDAPS predicts correct semantic and instance segmentation masks for the motor-bike (top-right) and the rider (botom-left), while EDAPS fails to do so. 

\noindent In summary, our contributions are as follows:
\vspace{-5px} 
\begin{enumerate}
    \item We introduce IMix, a novel target-to-source instance-aware cross-domain mixing strategy that generates exhaustively labeled source images with target instances for improved recognition quality {(i.e. reduced false positives and negatives)}.
    \item We reduce the catastrophic forgetting that arises when training with IMix by introducing CLIP-based domain alignment (CDA) as a regularizer.
    \item We combine both proposed modules to form LIDAPS, a language-guided instance-aware domain-adapted panoptic segmentation model that achieves SOTA results across multiple benchmarks.
\end{enumerate}
\noindent While we propose an end-to-end model with LIDAPS, our individual contributions remain orthogonal to the development of better panoptic UDA frameworks and are model-agnostic. Furthermore, both contributions can be detached during inference and thus do not induce any memory or computational constraints during inference.


\section{Related works}
 Panoptic segmentation~\cite{panopticDeepLab, Kirillov_2019_CVPR, Xiong_2019_CVPR, Li_2022_CVPR, Wang_2021_CVPR, Hu_2023_CVPR, ODISE, Chang_2020_ACCV} is becoming increasingly important with the rise of autonomous driving and AR/VR. \\
\noindent \textbf{Unsupervised Domain Adaptation (UDA)} panoptic segmentation trains on source domain labeled data and target domain unlabeled data. To achieve good performance on the unlabeled target images, these methods incorporate UDA techniques. A common one is self-training on pseudo-labels generated from targets images by a teacher network~\cite{PiPa, UDAPredClass, 2PCNet, focusTarget, IR2F, unidaformer, cvrn, fda} or the model itself~\cite{udapseudo,  kothandaraman2021ss}. Some works also refine and stabilize pseudo-labels~\cite{zhang2023black, IR2F, DIDA, zhang2021prototypical}.
Other approaches include adversarial strategies~\cite{hoffman2016fcns, luo2019taking, adverse, cycada, dlow, tsai2018learning, advent}, contrastive learning~\cite{PiPa, CONFETI, CDA-SAR}, regularizors~\cite{DAFast, DAP, tarvai, fixmatch, zhang2019category}, multiple resolution~\cite{hrda, hoyer2023domain} and domain adaptive architecture design~\cite{unidaformer,MemCDT}. In UDA works, only a few~\cite{cvrn, unidaformer, edaps} address panoptic segmentation. However, unlike existing work, we explore instance-aware cross-domain mixing to adapt the instance branch while simplifying the learning of difficult target objects by pasting them onto easy-to-segment source backgrounds. Furthermore, we are the first to exploit the domain-robustness of language-vision models to further align the source and target domains for panoptic UDA.


\noindent \textbf{Language in Segmentation} 
Several segmentation studies incorporate language to enhance their performance. This trend originated with DenseCLIP~\cite{DenseClip}, an extension of CLIP~\cite{clip} designed for dense downstream applications. While we also leverage dense per-pixel text similarity maps similar to DenseCLIP, as opposed to applying alignment on supervised images, we utilize the maps to align both the source and target domains via ground truth and generated pseudo-labels with domain-invariant CLIP text embeddings.
Importantly, unlike DenseCLIP which applies this knowledge distillation to the encoder features, we apply deep in the semantic decoder to prevent losing class-agnostic features in the shared encoder that are key for the task of instance segmentation. Open-vocabulary segmentation works also largely integrate language into their architecture~\cite{OVSeg, ODISE, ConvolutionsDieHard, SAN, ZUTIS, ViLD, xu2022simple, clip-fps}. These works do not perform unsupervised domain alignment. In some previous works~\cite{OVSegmentor, SFTextSuper, IFSeg, reclip}, mask annotations are unavailable, and large vision-language models are solely relied on for knowledge distillation. In contrast, we leverage the direct supervision available from a source domain. Some domain generalization segmentation works~\cite{ClipTheGAP, PTDiffSeg} also incorporate language to align their source embeddings with large language-vision embeddings to generalize to the target domain. However, these works~\cite{PTDiffSeg, ClipTheGAP} do not address instance segmentation which is specifically challenging given that CLIP mainly consists of semantic knowledge. Moreover, while some works~\cite{PADCLIP, videoUDA, adclip} investigate the incorporation of CLIP in UDA, only a few explore its effects in UDA segmentation. For instance, Chapman~\textit{et. al}~\cite{UDAPredClass} uses CLIP for UDA instance segmentation on an image level. In contrast, our work utilizes CLIP in a panoptic setting and calculates the text similarity on a pixel level. \\
\textbf{Augmented Data for Domain Adaptation} 
A key strategy in UDA segmentation involves training on augmented images. A common approach is the stylization and augmentation of images~\cite{cvrn, CONFETI, wildnet, melas2021pixmatch, hoyer2023mic, araslanov2021self} or the features of source images~\cite{liu2021learning, ClipTheGAP, poda}. Another approach is to leverage diffusion models and GANs to translate the style of source images or to synthesize training images~\cite{spigan, CDA-SAR, ControlUDA, DADiff, DIDEX, choi2019self, li2019bidirectional, wang2020differential, xpaste}. Zhao~\textit{et. al}\cite{xpaste} generates images of instances using generative models and then crops the instances using pre-trained segmentation models in order to do cross-domain mixing. This is not applied in a UDA setting and furthermore uses pre-trained segmentation models while we use our own EMA teacher network to predict the pseudo-masks of the target images for cropping. An alternative mainstream tactic is cross domain mix sampling (CDMS)~\cite{zhou2022context, dacs, cardace2022shallow}. ClassMix~\cite{classmix}, a CDMS technique, pastes pixels from half of the source image semantic classes onto a target image~\cite{dacs, IDA, HIMix, daformer, PRN, DAP}. Cardace~\textit{et. al}\cite{cardace2022shallow}, another CDMS technique,  pastes semantic masks from target to source. However, instance-aware mixing for the domain invariance enhancement of the instance decoder, specifically in panoptic segmentation models, remains largely unexplored. Lu~\textit{et. al}~\cite{instanceMix} explores instance mixing from source-to-target for UDA in action detection but neglects to refine the pseudo-masks. In contrast, we employ confidence-based thresholding to refine the pseudo-instance-masks which we find is key to reduce the confirmation bias. Furthermore, we apply the mixing in the opposite direction which yields a considerable performance gain by avoiding further bias injected due to an incomplete set of pseudo-labels arising from false negative predictions. 

\section{Method}\label{sec:method}

\begin{figure*}[tb]
  \centering
\includegraphics[width=\textwidth]{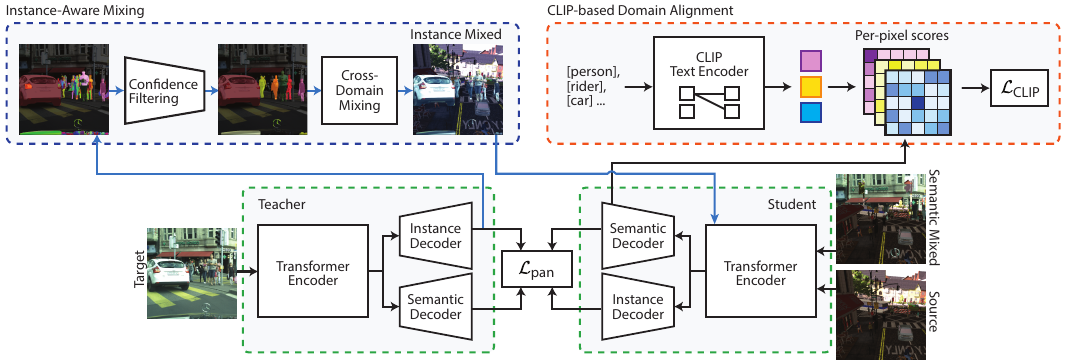}
  \caption{Illustration of the LIDAPS pipeline. (Green) The baseline panoptic UDA model is built on a mean-teacher framework and consists of a common transformer encoder and individual task decoders. The student model is supervised directly from source domain labels as well as semantically mixed inputs whose labels are generated by the teacher model. (Blue) We apply IMix to further adapt the instance segmentation branch of LIDAPS, mixing high-confidence predicted target instances with source images. Blue paths are only active when self-training with IMix is enabled. (Orange) We regularize the semantic branch via CLIP-based domain alignment that utilizes similarity maps to reduce catastrophic forgetting. 
  }
  \label{fig:pipeline}
\vspace{-14px}\end{figure*}

\subsection{Preliminary}\label{sec:preliminary}

\textbf{Panoptic segmentation} is commonly tackled by breaking it down into its subtasks: semantic and instance segmentation. A panoptic segmentation model is thereby trained on a panoptic segmentation loss $\mathcal{L}_\mathrm{pan}$
given by the sum of a semantic and an instance loss:
\vspace{-2px}
\begin{equation} 
    \mathcal{L}_\mathrm{pan} = \mathcal{L}_\sem + \mathcal{L}_\mathrm{inst}
\label{eq:main-inst-sem}
\end{equation}
\vspace{-3px}
In this work, for semantic segmentation, we use pixel-wise categorical cross-entropy loss,
while for instance segmentation, we follow a top-down approach,
and compute RPN and RoIAlign box regression and classification losses following MaskRCNN~\cite{maskrcnn}.
\noindent \textbf{Panoptic UDA} is the task of transferring knowledge from a learned source domain to a target domain. In this setup, a machine learning model $\phi$ is trained on both source  $\mathcal{D}^{s}=\{x_i^{s}, y_i^{s}\}_{i=1}^{N^{s}}$ and target domain images $\mathcal{D}^{t}=\{x_i^{t}\}_{i=1}^{N^{t}}$, with direct human annotated supervision only available on the source domain via semantic $y^{s}_{{\sem}} \in \mathbb{R}^{H \times W \times C}$ and instance labels $y^{s}_{{\textrm{inst}}} \in \mathbb{R}^{H \times W \times B}$. Here $C$ denotes the number of semantic classes and $B$ denotes the number of ground truth instances given images of size $H \times W$.
The na\"ive approach to tackling panoptic UDA is to treat the problem similar to standard supervised training, and thus only train with the supervision from the source labels via the source loss $\mathcal{L}^{s}$. However, the variance in data distribution between the source and target images, i.e. the ``domain gap'', severely limits the transferability of learned knowledge across domains. Such a na\"ive approach consisting of only a source loss thus remains inadequate for achieving a good performance on the target domain.

Self-training is a common technique used to reduce the domain gap between source and target by leveraging a model's own predictions to extend the supervision to the target domain~\cite{zou2018unsupervised,zhang2019category,mei2020instance,dacs,zhang2021prototypical,hoyer2021improving, daformer, edaps}. In this work, we adopt a self-training approach that entails both the supervised loss on the source domain $\mathcal{L}^{s}_{pan}$, and a self-supervised loss $\mathcal{L}^{ss}_{pan}$, resulting in the final training objective:
\vspace{-2px} 
\begin{equation}
    \argmin_\phi \, \mathcal{L}^{s}_{pan} + \mathcal{L}^{ss}_{pan}
\end{equation}
\vspace{-3px} 
\subsection{Establishing a Baseline for Panoptic UDA} \label{sec:naive_baseline}

In a self-training framework, a model learns from its own predictions. This however can result in \textit{confirmation bias} as the model trains on incorrect pseudo-labels, therefore commonly, predictions are refined prior to application~\cite{IR2F, DIDA}. The mean-teacher framework~\cite{tarvai} proposes a simple but effective way to generate stabilized on-the-fly pseudo-labels by leveraging the fact that the stochastic averaging of a model's
weights yields a more accurate model than using the final training weights directly. A mean-teacher framework is therefore built with two models, namely the student that is trained (e.g. via gradient decent), and the teacher $\theta$ whose weights are updated based on the exponential moving average (EMA) of successive student weights:
\vspace{-2.0mm} 
\begin{equation}
    \phi_{t+1} \leftarrow \alpha \phi_t + (1 - \alpha) \theta_t\,\footnote{We use the notation for the model and its weights interchangeably for readability.}
\end{equation}
for time step $t$ and $\alpha$ that denotes the smoothing coefficient.
\vspace{-0.2mm} 
While the mean-teacher extends the supervision to the target domain, the supervisory signal remains highly noisy and may still destabilize the training process. A common solution applied in UDA setups is to employ cross-domain mixing to generate images that contain both noisy target domain information and clean source domain ground truth annotations~\cite{dacs, IDA, HIMix, daformer, PRN, DAP}. Specifically, the teacher network $\theta$ predicts the pseudo-labels for the target image that forms the augmented input for the student via a cut-and-paste operation on the source image. Formally, given a binary mask of semantic labels to be cut $\mathbf{M}_{\textrm{sem}} \in \{0,1\}^{H \times W}$ and target semantic pseudo-labels generated by the teacher model $y^t_{\sem}$, the semantic cross-domain mixed sampling (DACS) can be defined as:
\vspace{-1.0mm} 
\begin{equation} \label{eq:sem_mix}  
\begin{split}
    \tilde{x} &= \mathbf{M}_{\textrm{sem}} \odot x^s + (1 - \mathbf{M}_{\textrm{sem}}) \odot x^t \\
    \tilde{y}_{\textrm{sem}} &= \mathbf{M}_{\textrm{sem}} \odot y_{\textrm{sem}}^s + (1 - \mathbf{M}_{\textrm{sem}}) \odot y_{\textrm{sem}}^t \\
\end{split}
\end{equation}
with $\odot$ denoting a dot product and $\tilde{\cdot}$ indicating the mixed domain. Such DACS operations leveraging ClassMix~\cite{classmix} coupled with self-training have shown significant performance gains when tackling semantic UDA~\cite{daformer}, with the core idea stemming from consistency regularization~\cite{classmix, sajjadi2016regularization, tarvai, fixmatch} which states that predictions for unlabelled data should be invariant to perturbations or augmentation.
\vspace{-1.0mm}
\noindent Firstly, the supervised semantic loss consists of :
\begin{equation} \label{eq:sem-source}
\vspace{-1.0mm}
     \mathcal{L}^{s}_\textrm{sem}(\hat{y}_\textrm{sem}^{s}, y_{\textrm{sem}}^{s}) = -\sum_{i,j,c} \left(y_\textrm{sem}^{s} \log(\hat{y}_\textrm{sem}^{s})\right)_{i,j,c}
 \end{equation}
 \vspace{-1.0mm}

Formally, the self-supervised loss for the semantically adapted self-training baseline, built on a weighted cross-entropy, is given by:
\vspace{-1.0mm}
\begin{equation}
\label{eq:baseline_loss}
  \mathcal{L}^{ss}_{pan} =  \mathcal{L}^{ss}_{sem}(\hat{\tilde{y}}_{\textrm{sem}}, \tilde{y}_{\textrm{sem}})
\end{equation}
with $\hat{\cdot}$ denoting the prediction of the model and 
\vspace{-1.0mm}
\begin{equation}
  \mathcal{L}^{ss}_{sem}(\hat{\tilde{y}}_{\textrm{sem}}, \tilde{y}_{\textrm{sem}}) = 
  \begin{cases}
    \mathcal{L}^s_{\sem}(\hat{\tilde{y}}_\sem, y^s_\sem), \\
    \hphantom{ozan} \textrm{ if } \mathbf{M}_{\textrm{sem}}^{(h,w,c)} = 1, \\
    -\sum
    k^t_{(h,w)}
    \left( 
      y^t_{\sem} \log(\hat{\tilde{y}}_{\sem})
    \right)_{(h,w,c)}, \\
    \hphantom{ozan} \textrm{ otherwise} \\
    \end{cases}
\end{equation}
\vspace{-1.0mm}
with $k^t$ the model's estimated confidence for the pseudo-label~\cite{dacs}. Specifically, we apply a cross-entropy loss on pixels coming from the source image ($\mathbf{M}_{\textrm{sem}}^{(h,w,c)} = 1$), and apply a weighted loss on the pixels coming from the target image, supervised via the teacher generated pseudo-label $y^t_{\sem}$.
We illustrate this baseline in Fig.~\ref{fig:pipeline} - green.

Given that semantic segmentation forms one-half of panoptic segmentation, such a baseline approach that adapts the semantic maps between the source and target domains via DACS can significantly reduce the domain gap for panoptic segmentation. However, such an approach forgoes a crucial element of panoptic UDA altogether, adapting the instance segmentation task between two domains. In fact, DACS does not generate
augmented images containing sufficient instance-specific information to adapt the instance branch.
In the following section, we tackle the adaptation of instance segmentation between a source and target domain to improve panoptic segmentation performance. 

\subsection{Instance-Aware Mixing (IMix)}\label{sec:imix}

\begin{figure*}[t]
    \centering
    \begin{subfigure}{.24\textwidth}
        \centering
        \includegraphics[width=\linewidth]{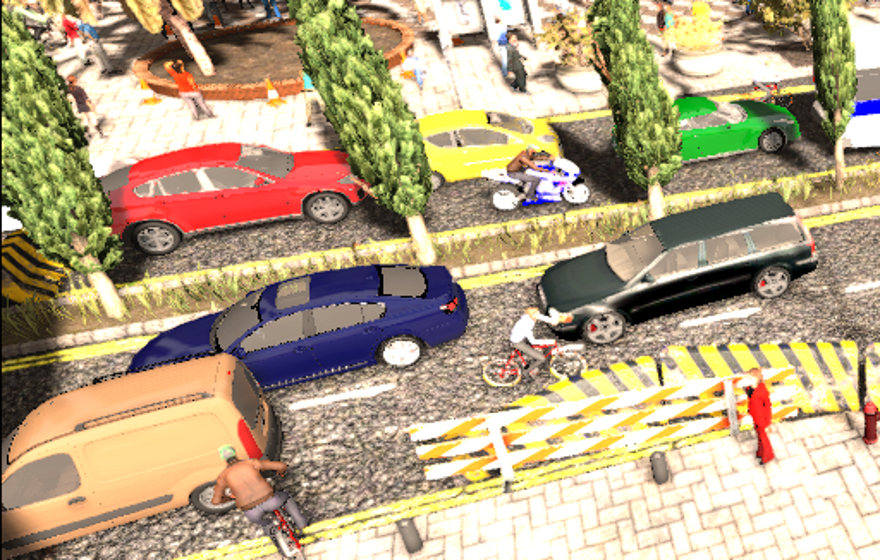}
        \caption{Source Image}
        \label{fig:imix_a} 
    \end{subfigure}\hfill
    \begin{subfigure}{.24\textwidth}
        \centering
        \includegraphics[width=\linewidth]{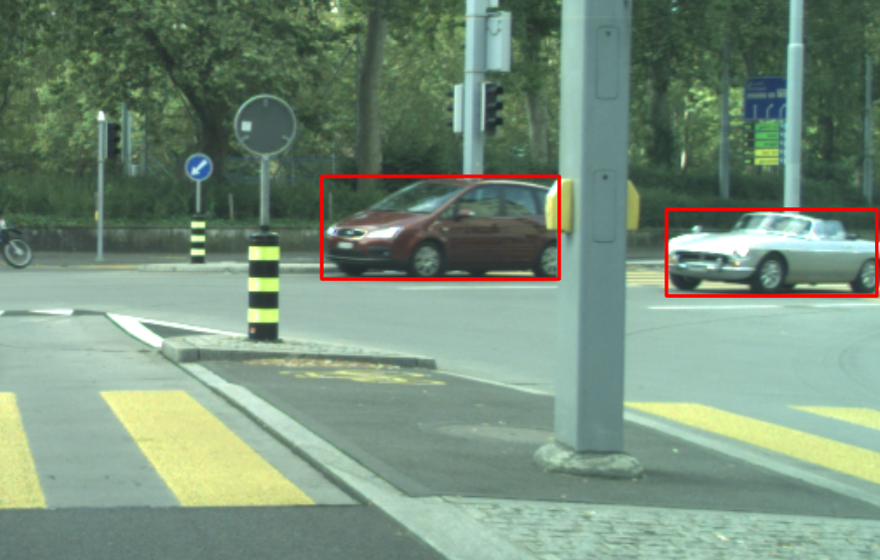}
        \caption{Target Image}
        \label{fig:imix_b} 
    \end{subfigure}\hfill
    \begin{subfigure}{.24\textwidth}
        \centering
        \includegraphics[width=\linewidth]{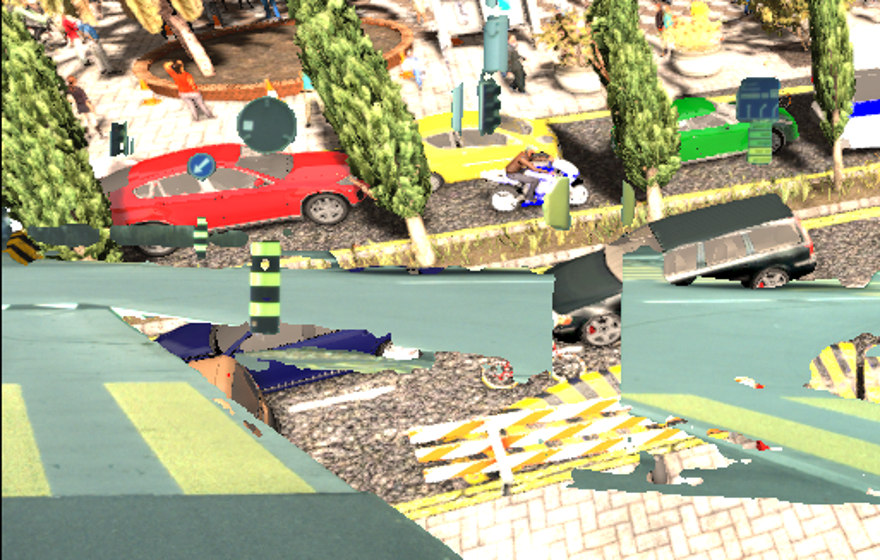}
        \caption{DACS}
        \label{fig:imix_c} 
    \end{subfigure}\hfill
    \begin{subfigure}{.24\textwidth}
        \centering
        \includegraphics[width=\linewidth]{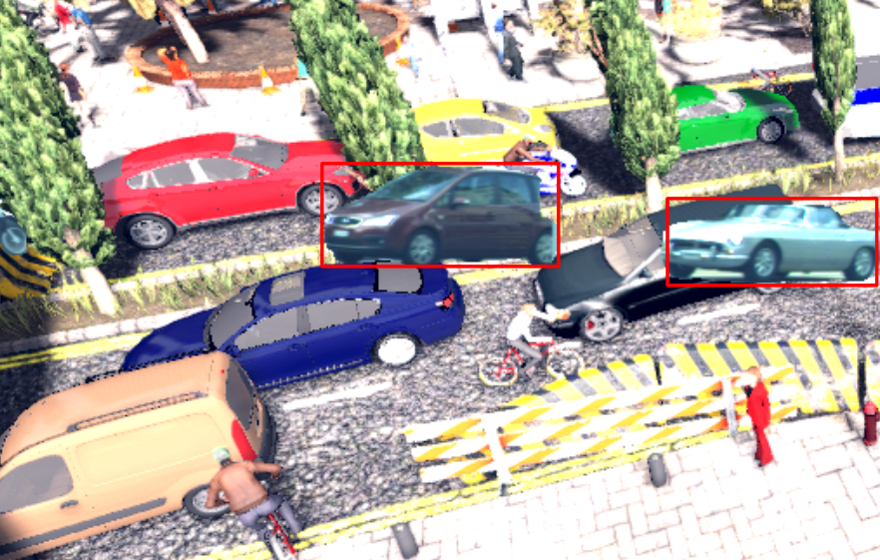}
        \caption{IMix (ours)}
        \label{fig:imix_d} 
    \end{subfigure}  
    \caption{Our Proposed IMix cuts and pastes pseudo-instances with a confidence score above a certain threshold from source to target while in DACS, half of the semantic classes are pasted from source to target without preserving instance-level information.}
    \label{fig:imix_all}
\vspace{-15px} \end{figure*}

We propose a novel mixing strategy called IMix, to reduce the domain gap when tackling instance segmentation. The goal of IMix is to apply cross-domain mixed sampling while not only retaining instance-level information but also simplifying the recognition of target objects by presenting them within source environments. A sample image utilizing IMix is compared to DACS in Fig.~\ref{fig:imix_all}.

However, mixing source and target domain information on an instance level raises a crucial challenge, stemming from how the two tasks are supervised. Unlike semantic segmentation where losses are applied on a pixel level, instance segmentation is typically supervised by the injective function that maps the set of ground truth objects to the set of predicted instances. Therefore an instance segmentation model remains prone to confirmation bias if ground truth label exhaustiveness is not guaranteed, i.e. the model will learn to incorrectly identify objects as background if every visible object within the scene does not have an associated instance mask (see supp. for false negative example).

A mixing operation must account for such a challenge. We thus construct IMix such that it is handled from target-to-source, avoiding the incompleteness of instance labels that may emerge from false negative predictions. Formally: 
\vspace{-7px}
\begin{equation} \label{eq:inst_mix}
\begin{split}
    \tilde{x} &= \mathbf{M}_{\textrm{inst}} \odot x^t + (1 - \mathbf{M}_{\textrm{inst}}) \odot x^s \\
    \tilde{y}_{\textrm{inst}} &= \mathbf{M}_{\textrm{inst}} \odot y_{\textrm{inst}}^t+ (1 - \mathbf{M}_{\textrm{inst}}) \odot y_{\textrm{inst}}^s \\
\end{split}
\end{equation}
with $y_{\textrm{inst}}^t$ and  $y_{\textrm{inst}}^s$ denoting the target pseudo-label and source ground truth label respectively, and $\mathbf{M}_{\textrm{inst}} \in \{0,1\}^{H \times W}$ the sum of binary instance masks based on the teacher's prediction.

Specifically, we cut the instances from the teacher model's output and paste them onto a source image, constructing the mixed pseudo-label by merging the ground truth instance labels with the teacher's predictions. Utilizing the source image as a background ensures all visible objects have corresponding mask annotations. Furthermore, since a model learns a source domain much more efficiently thanks to direct supervision, with IMix, we simplify the recognition task of target instances by presenting them on easy-to-separate source domain environments.

However, while retaining exhaustiveness limits confirmation bias caused by false negative predictions, a self-supervised model is still prone to such effects due to false positives. In other words, if incorrect instance masks are pasted on the mixed image, the model will learn to affirm its preexisting biases, causing an increased number of false positive predictions. To minimize such cases, we propose a simple but effective confidence filtering step. We predict a confidence score alongside the instance masks of each object~\cite{maskrcnn}. We apply filtering based on the predicted confidence values to redefine the joint mixing mask as:
\vspace{-2px}
\begin{equation}
    \mathbf{M}_{\textrm{inst}} = \sum_{i \in I} \mathbbm{1}[h_i^t > \tau] \, y^t_{\textrm{inst},i}
\end{equation}
\vspace{-2px}
with $I$ denoting the set of predicted instances, $y^t_{\textrm{inst},i}$ the predicted $i$'th instance mask, $h_i^t$ the corresponding confidence score and $\tau$ the threshold hyperparameter. Thus, our self-supervised panoptic loss from Eq.~\ref{eq:baseline_loss} can be updated as:
\begin{equation} \label{eq:updated_loss}
  \mathcal{L}^{ss}_{pan} = \mathcal{L}^{ss}_{sem}(\hat{\tilde{y}}_{\textrm{sem}}, \tilde{y}_{\textrm{sem}}) + \mathcal{L}^{ss}_{\textrm{inst}}(\hat{\tilde{y}}_{\textrm{inst}}, {\tilde{y}_{\textrm{inst}}})
\end{equation}

As commonly seen in multitask frameworks, the increase in supervisory signals from one task may cause catastrophic forgetting for another, i.e. the weights in the network that are important for one task may be changed to meet the objectives of another~\cite{PADCLIP}. We observe similar behavior in our training when fine-tuning LIDAPS on IMix (please refer to Sec.~\ref{sec:exp}). Specifically, the performance gains of our model for panoptic segmentation are hindered by the drop in semantic quality. In the following section, we address this problem by introducing a language-based regularization for semantic segmentation.
\vspace{-1.5px}
\subsection{CLIP-based Domain Alignment (CDA)}\label{sec:cda}
A simple but effective solution to reducing catastrophic forgetting when multitask learning is to leverage the embedding space of a pre-trained model as an anchor, i.e. the intermediate features as continual auxiliary targets, which is also commonly employed in unsupervised domain adaptation frameworks to limit overfitting onto the source domain~\cite{daformer, edaps}. In this work we exploit both use cases for weight anchoring by relying on CLIP~\cite{clip} embeddings to regularize the semantic branch of our network. CLIP is trained on a very large-scale image-text pair dataset, providing a diversified, robust world model. We argue that by semantically aligning each domain to the CLIP embedding space, we can implicitly enforce domain invariance. In other words, we train our model such that the features of a source or target image both aim to generate high similarities to a joint CLIP embedding. An illustration of CLIP-based domain alignment (CDA) can be seen in Fig.~\ref{fig:pipeline} - orange.

\begin{figure*}[t]
    \centering
    \includegraphics[width=0.9\textwidth]{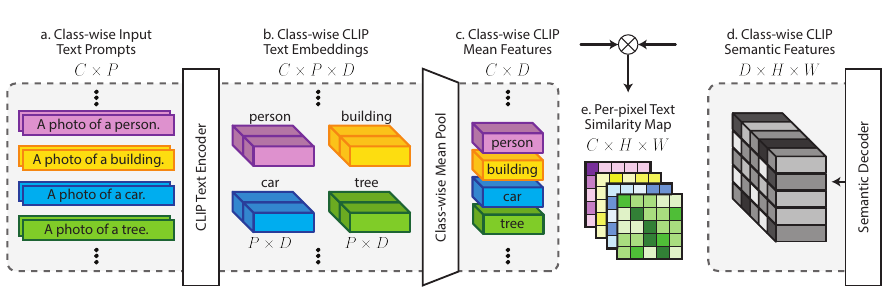}
    \vspace{-8px} 
    \caption{Pipeline used to compute the pixel-text similarity map for CLIP-based domain alignment. We generate class-wise CLIP mean features from a series of fixed text prompts (a-c). \textit{C}: $\#$semantic classes, \textit{D}: dimension of text encodings, \textit{H}, \textit{W}: width and height of image, \textit{P}: $\#$prompts (e.g. A painting of a person). The similarity maps are given by the inner product with the semantic decoder features.}
    \label{fig:clip_align}
\vspace{-5px} \end{figure*}

However, to be able to exploit CLIP features to avoid the divergence of semantic features of source and target images, regularization needs to be applied deep within the network. This of course imposes limitations on the expressibility of the features or effectiveness of the regularization when directly using CLIP embeddings as targets. To this end, we construct a pixel-level representation from natural language prompts following DenseCLIP~\cite{DenseClip} and only supervise the similarity to the semantic decoder features. Specifically, our CLIP-based domain alignment strategy follows two steps as illustrated in Fig.~\ref{fig:clip_align}. We first generate class-wise mean CLIP features by mean pooling over the CLIP embeddings generated from $P$ text prompts for $C$ semantic classes following set precedent~\cite{DenseClip}, with $P$ denoting the number of text prompts per class (Fig.~\ref{fig:clip_align}a-c). Each row in the resulting matrix represents a CLIP embedding that encodes meaningful semantic information about a particular class. These embeddings act as anchors within our alignment module, with each generated semantic feature (Fig.~\ref{fig:clip_align}d) aiming to achieve high similarity with a semantically corresponding vector. Finally, we compute the per-pixel text similarity maps $\sigma^{\text{sim}}$ through the inner product of the decoder features and mean CLIP features (Fig.~\ref{fig:clip_align}e).

Formally, the CLIP-based domain alignment loss can be stated as follows:
\vspace{-7px}
\begin{equation} \label{eq:clip-align-loss}
    \mathcal{L}_{\text{CLIP}} = -\frac{1}{HWC} \sum_{h=1}^{H} \sum_{w=1}^{W} \sum_{c=1}^{C} \mathbbm{1}[y_{(h,w)} = c] \log\left(\hat{y}_{(h,w,c)}^{\text{sim}}\right),
\end{equation}
with $\mathbbm{1}[\cdot]$ denoting the indicator function and $\hat{y}^{\text{sim}}$ denoting the embedding-text similarity probability given by:
\vspace{-1mm} 
\begin{equation}
\hat{y}_{(h,w,c)}^{\text{sim}} =  \frac{\exp\left(\sigma_{(h,w,c)}^{\text{sim}}\right)}{\sum_{c'=1}^{C} \exp\left(\sigma_{(h,w,c')}^{\text{sim}}\right)}.
\end{equation}
\vspace{-1mm}
\noindent In our proposed LIDAPS model, the CLIP loss is incorporated in our $\mathcal{L}_{sem}$ loss with a weight loss of 1.0.
\vspace{-1.0px}

\section{Experiments} \label{sec:exp}

We follow standard evaluation protocols and datasets for panoptic UDA following set precedent~\cite{cvrn, unidaformer, edaps}. All reported values are the averaged scores over three runs with three different seeds (1, 2, 3). For further details including the implementation details including architecture and hyper-parameter details please see the supplementary.

\subsection{Results}
We compare our proposed LIDAPS with other state-of-the-art (SOTA) UDA panoptic segmentation methods on four different benchmarks including SYNTHIA $\rightarrow$ Cityscapes (S$\rightarrow$C),  SYNTHIA $\rightarrow$ Mapillary Vistas (S$\rightarrow$M), Cityscapes $\rightarrow$ Mapillary Vistas (C$\rightarrow$M) and Cityscapes $\rightarrow$ Foggy Cityscapes (C$\rightarrow$F). As seen in Tab.~\ref{TableBenchmarks}, our model consistently outperforms existing works across the board, exceeding the performance of previous SOTA by up to +3.6 mPQ.
Furthermore in Fig.~\ref{fig:results}, we provide qualitative results demonstrating the capabilities of LIDAPS. Compared to EDAPS~\cite{edaps}, LIDAPS can better separate semantically similar neighboring instances by leveraging instance-aware adaptation via IMix and retain its semantic quality via CDA.
\newcolumntype{R}[2]{%
    >{\adjustbox{angle=#1,lap=\width-(#2)}\bgroup}%
    l%
    <{\egroup}%
}
\newcommand*\rot{\multicolumn{1}{R{45}{1em}}}
\begin{table*}[t]
 \centering
 \caption{Class-wise comparison to SOTA on four different benchmarks for UDA panoptic segmentation. Reported results are averaged over three runs with three different seeds. }
 \vspace{-2.5mm}
 \scalebox{.66}{
 \vspace{-3.5mm} 
\begin{tabular}{ l | c c c c c c c c c c c c c c c c | c c c }

Method & \rot{road} & \rot{sidewalk} & \rot{building} & \rot{wall} & \rot{fence} & \rot{pole} & \rot{light} & \rot{sign} & \rot{veg} & \rot{sky} & \rot{person} & \rot{rider}& \rot{car}& \rot{bus}&\rot{m.bike} & \rot{bike} & $\mathrm{mSQ}$ & $\mathrm{mRQ}$ & $\mathrm{mPQ}$\\
\hline
\hline
    \multicolumn{20}{c}{SYNTHIA $\rightarrow$ Cityscapes}
\\
\hline
FDA\cite{fda} & 79.0 & 22.0 & 61.8 & 1.1 & 0.0 & 5.6 & 5.5 & 9.5 & 51.6 & 70.7 & 23.4 & 16.3 & 34.1 & 31.0 & 5.2 & 8.8 & 65.0 & 35.5 & 26.6 \\
CRST\cite{udapseudo} & 75.4 & 19.0 & 70.8 & 1.4 & 0.0 & 7.3 & 0.0 & 5.2 & 74.1 & 69.2 & 23.7 & 19.9 & 33.4 & 26.6 & 2.4 & 4.8 & 60.3 & 35.6 & 27.1 \\
AdvEnt\cite{advent} & \textbf{87.1} & 32.4 & 69.7 & 1.1 & 0.0 & 3.8 & 0.7 & 2.3 & 71.7 & 72.0 & 28.2 & 17.7 & 31.0 & 21.1 & 6.3 & 4.9 & 65.6 & 36.3 & 28.1 \\
  CVRN\cite{cvrn} &86.6 & 33.8 & 74.6 & 3.4 & 0.0 & 10.0 & 5.7 & 13.5 & 80.3 & 76.3 & 26.0 & 18.0 & 34.1 & 37.4 & 7.3 & 6.2 & 66.6 & 40.9 & 32.1\\
  UniDAformer\cite{unidaformer} &73.7 & 26.5 & 71.9 &1.0 & 0.0&7.6 &9.9 & 12.4&81.4 &77.4 &27.4 & 23.1&47.0 & \textbf{40.9} &12.6 & \textbf{15.4} &64.7 &42.2 & 33.0\\
    EDAPS\cite{edaps} & 77.5 & 36.9 & 80.1 & 17.2 & 1.8 & 29.2 & 33.5 & 40.9 & 82.6 & 80.4 & 43.5 & 33.8 &45.6 & 35.6 & 18.0 & 2.8 & 72.7 & 53.6 & 41.2 \\ \hline
 LIDAPS(ours) & 80.8 & \textbf{48.8} & \textbf{80.8} & \textbf{17.6} & \textbf{2.5} & \textbf{29.9} & \textbf{34.6} & \textbf{42.9} & \textbf{82.8} & \textbf{82.9} & \textbf{44.4} & \textbf{40.5} & \textbf{51.7} & 39.2 & \textbf{27.4} & 10.7 & \textbf{74.4} & \textbf{57.6} & \textbf{44.8}\\ 
\hline
\hline
    \multicolumn{20}{c}{Cityscapes $\rightarrow$ Foggy Cityscapes}
 \\
\hline
DAF\cite{DAFast}& \textbf{94.0} &54.5& 57.7 &6.7 &10.0& 7.0& 6.6 &25.5& 44.6& 59.1 &26.7& 16.7 &42.2& 36.6 &4.5& 16.9& 70.6 &41.7 &31.8\\
FDA\cite{fda} & 93.8& 53.1 &62.2 &8.2& 13.4 &7.3& 7.6 &28.9& 50.8 &49.7 &25.0 &22.6& 42.9 &36.3 &10.3& 15.2 &71.4& 43.5 &33.0\\
AdvEnt\cite{advent} & 93.8 &52.7& 56.3& 5.7& 13.5& 10.0& 10.9& 27.7& 40.7& 57.9& 27.8& 29.4& 44.7& 28.6 &11.6& 20.8& 72.3& 43.7 &33.3\\
CRST\cite{udapseudo} & 91.8& 49.7& 66.1 &6.4& 14.5 &5.2& 8.6& 21.5& 56.3& 50.7& 30.5& 30.7& 46.3& 34.2 &11.7& 22.1& 72.2& 44.9 &34.1\\
SVMin\cite{svmin} & 93.4 & 53.4 & 62.2 & 12.3 & 15.5 & 7.0 & 8.5 & 18.0 & 54.3 & 57.1 & 31.2 & 29.6 & 45.2& 35.6 &11.5& 22.7 &72.4 &45.5 &34.8 \\
 CVRN\cite{cvrn} & 93.6 & 52.3 & 65.3 & 7.5 & 15.9 & 5.2 & 7.4 & 22.3 & 57.8 & 48.7 & 32.9 & 30.9 & 49.6 & 38.9 & 18.0 & 25.2 & 72.7 & 46.7 & 35.7\\
 UniDAformer\cite{unidaformer} & 93.9 & 53.1 & 63.9 & 8.7 & 14.0 & 3.8 & 10.0 & 26.0 & 53.5 & 49.6 & 38.0 & 35.4 & 57.5 & 44.2 & 28.9 & 29.8 & 72.9 & 49.5 & 37.6\\
  EDAPS\cite{edaps} & 91.0 & 68.5 & 80.9 & \textbf{24.1} & 29.0 & 50.1 & 47.2 & 67.0 & 85.3 & 71.8 & 50.9 & 51.2 & 64.7 & 47.7 & 36.9 & 41.5 & 79.2  & 70.5  & 56.7\\ \hline
 LIDAPS(ours) & 92.3 & \textbf{70.0} & \textbf{83.2} & 23.8 & \textbf{31.9} & \textbf{56.4} & \textbf{47.7} & \textbf{68.8} & \textbf{86.6} & \textbf{72.5} & \textbf{53.2} & \textbf{53.6} & \textbf{68.0} & \textbf{56.6} & \textbf{42.8} & \textbf{45.9} &\textbf{80.2} & \textbf{73.2} & \textbf{59.6} \\ 
 \hline
 \hline
     \multicolumn{20}{c}{SYNTHIA $\rightarrow$ Mapillary Vistas}
\\
\hline
FDA\cite{fda} & 44.1 & 7.1 & 26.6 & 1.3 & 0.0 & 3.2 & 0.2 & 5.5 & 45.2 & 61.3 & 30.1 & 13.9 & 39.4 & 12.1 & 8.5 & 7.0 & 63.8 & 26.1 & 19.1 \\
CRST\cite{udapseudo} & 36.0 & 6.4 & 29.1 & 0.2 & 0.0 & 2.8 & 0.5 & 4.6 & 47.7 & 68.9 & 28.3 & 13.0 & 42.4 & 13.6 & 5.1 & 2.0 & 63.9 & 25.2 & 18.8 \\
AdvEnt\cite{advent} & 27.7 & 6.1 & 28.1 & 0.3 & 0.0 & 3.4 & 1.6 & 5.2 & 48.1 & 66.5 & 28.4 & 13.4 & 40.5 & 14.6 & 5.2 & 3.3 & 63.6 & 24.7 & 18.3 \\
 CVRN\cite{cvrn} &33.4 & 7.4 & 32.9 & 1.6 & 0.0 & 4.3 & 0.4 & 6.5 & 50.8 & 76.8 & 30.6 & 15.2 & 44.8 & 18.8 &7.9 & \textbf{9.5} &65.3 & 28.1 & 21.3\\
  EDAPS\cite{edaps} & \textbf{77.5} & \textbf{25.3} & 59.9 & \textbf{14.9} & 0.0 & 27.5 & 33.1 & \textbf{37.1} & \textbf{72.6} & 92.2 & 32.9 & 16.4 & 47.5 & 31.4 & 13.9 & 3.7 & 71.7  & 46.1  & 36.6\\ \hline
 LIDAPS(ours) & 76.5& 25.2 & \textbf{64.2} & 14.0 & \textbf{0.2} & \textbf{29.1} & \textbf{35.6} & 35.3 & 72.1 & \textbf{94.4} & \textbf{33.8} & \textbf{18.3} & \textbf{50.3} & \textbf{33.9} & \textbf{19.3} & 5.9 &\textbf{73.9} & \textbf{47.7} & \textbf{38.0} \\ 
\hline
\hline
 \multicolumn{20}{c}{Cityscapes $\rightarrow$ Mapillary Vistas}
\\
\hline
CRST\cite{udapseudo} & 77.0 & 22.6 & 40.2 & 7.8 & 10.5 & 5.5 & 11.3 & 21.8& 56.5& 77.6 & 29.4 & 18.4 & 56.0 & 27.7 & 11.9 & 18.4 & 72.4 & 39.9 & 30.8\\
FDA\cite{fda} & 74.3 & 23.4 & 42.3 & 9.6 & 11.2 & 6.4 & 15.4 & 23.5 & 60.4 & 78.5 & 33.9 & 19.9 & 52.9 & 8.4 & 17.5 & 16.0 & 72.3 & 40.3 & 30.9\\
AdvEnt\cite{advent}& 76.2 & 20.5 & 42.6 & 6.8 & 9.4 & 4.6 & 12.7 & 24.1 & 59.9 & 83.1 & 34.1 & 22.9 & 54.1 & 16.0 & 13.5 & 18.6 & 72.7 & 40.3 & 31.2 \\
 CVRN\cite{cvrn} & \textbf{77.3} & 21.0 & 47.8 & 10.5 & 13.4 & 7.5 & 14.1 & 25.1 & 62.1 & \textbf{86.4} & \textbf{37.7} & 20.4 & 55.0 & 21.7 & 14.3 & 21.4 & 73.8 & 42.8 & 33.5\\
  EDAPS\cite{edaps} & 58.8 & 43.4 & 57.1 & 25.6 & 29.1 & 34.3 & 35.5 & 41.2 & 77.8 & 59.1 & 35.0 & 23.8 & 56.7 & 36.0 & 24.3 & 25.5 & 75.9  & 53.4  & 41.2\\ \hline
 LIDAPS(ours) & 49.1 & \textbf{44.3} & \textbf{70.1} & \textbf{26.5} & \textbf{29.9} & \textbf{37.4} & \textbf{37.2} & \textbf{43.2} & \textbf{80.0}  & 46.1 & 35.9 & \textbf{25.0} & \textbf{57.1} & \textbf{41.6} & \textbf{29.6} & \textbf{28.4} & \textbf{76.6}& \textbf{54.9} &\textbf{42.6} 
\label{TableBenchmarks}
\end{tabular}
}
\vspace{-0px}\end{table*}

\begin{figure}[t]
    \centering
    \includegraphics[width=\columnwidth]{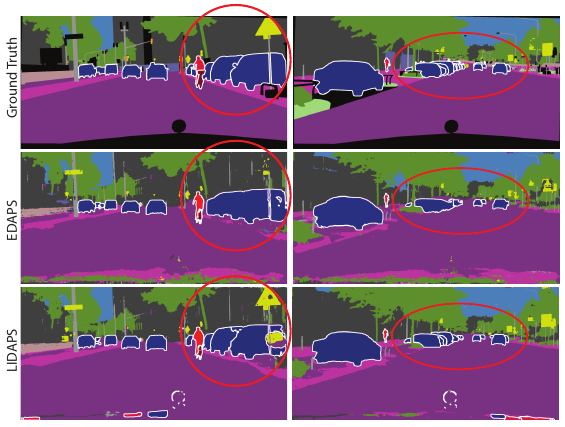}
    \caption{Qualitative results from SYNTHIA $\rightarrow$ Cityscape comparing EDAPS~\cite{edaps} to our proposed LIDAPS.}
    \label{fig:results}
\vspace{-8px}
\end{figure}

\begin{table}[t]
    \centering
    \begin{minipage}{0.48\textwidth}
        \centering
        \caption{Ablation study on proposed modules. Starting from a baseline EDAPS*, we introduce our instance-aware cross-domain mixing (IMix) and CLIP-based domain alignment (CDA).}
        \scalebox{0.83}{ 
            \begin{tabular}{ccc|ccc|cc}
                EDAPS$^*$ & IMix & CDA & mSQ & mRQ & mPQ & mIoU & mAP \\
                \hline
                \ding{51}  &  &   &  72.3 &53.3  & 41.0 &58.0  & 34.1 \\
                \ding{51}  & \ding{51} & & 73.0 & 54.7 & 42.3 &57.7 & 39.5  \\
                \ding{51}  &  &\ding{51} &  73.9 & 55.0 & 42.9 & 59.6 & 34.4  \\
                \ding{51} &  \ding{51}  & \ding{51} & \textbf{74.4} & \textbf{57.6} & \textbf{44.8} & \textbf{59.6} & \textbf{42.6} \\
            \end{tabular}
        }
        \label{TableAblateAll}
    \end{minipage}
    \vspace{-8px}
\end{table}
\begin{table}[t]
    \centering
    \tabcolsep=5pt
        \centering
        \caption{Ablation study on mixing strategy for panoptic segmentation comparing (i) the mixing direction when applying IMix, (ii) the effects of ClassMix\cite{classmix} when applied from target-to-source. S stands for source and T stands for target. The baseline is EDAPS*+CDA.}
        \scalebox{0.8}{ 
            \begin{tabular}{ll|cc|ccc|cc}
                & Method & Copy & Paste & mSQ & mRQ & mPQ & mIoU & mAP \\
                \hline
                & Baseline & - & - & 73.9 & 55.0 & 42.9 & 59.6 & 34.4\\
                \hline
                \multirow{2}{*}{(i)}  & \multirow{2}{*}{+ IMix} & S & T & 62.0 & 37.6 & 29.3 & 56.2 & 1.9\\
                & & T & S & \textbf{74.4} & \textbf{57.6} & \textbf{44.8} & \textbf{59.6} & \textbf{42.6}\\
                \hline
                (ii) & + ClassMix & T & S & 73.5& 53.9& 42.1 & 58.6 & 34.8\\
            \end{tabular}
        }
        \label{TablePasteDirection}
\vspace{-8px}
\end{table}

\subsection{Ablation Studies} \label{ablation-studies}
In this section, we conduct ablation studies on SYNTHIA $\rightarrow$ Cityscapes to demonstrate the effectiveness of our proposed components. The corresponding tables with the standard deviations alongside a hyper-parameter study for the confidence score threshold is found in the supplement.\\
\noindent \textbf{Effects of Network Components}: In Tab.~\ref{TableAblateAll}, we isolate the effects of the different modules of our proposed pipeline. Starting from the baseline EDAPS*, we introduce our cross-domain instance mixing (IMix) which improves the panoptic segmentation performance ($+1.3\%$ mPQ) through significant instance segmentation gain ($+5.4\%$ mAP). Due to their contradictory goals, the improvement in instance segmentation comes in lieu of semantic performance ($-0.3\%$ mIoU) that becomes subject to catastrophic interference. To remedy this we propose CLIP-based domain alignment (CDA). First, we separately introduce CDA to understand its isolated effects. As observed, the module aids the panoptic segmentation performance with a $+1.9\%$ mPQ improvement, which solely stems from the gains in semantic segmentation ($+1.6\%$ mIoU increase as opposed to the relatively unchanged mAP). Next, we showcase the combined effects of the two components which improve both the semantic and instance segmentation performance of our baseline, allowing LIDAPS to achieve $44.8\%$ mPQ. As seen, the final model demonstrates significant gains in instance segmentation ($+8.5$ mAP) thanks to IMix, while retaining its semantic segmentation gains from the CLIP-based domain alignment ($+1.6\%$ mIoU). CDA alone inscreases the final mPQ more than IMix alone because semantic seg. includes 16 classes (both ``thing'' and ``stuff'' classes), while instance seg. considers only 6 thing (countable) classes where ``stuff'' classes (e.g. sky) occupy a significant number of pixels. 

\noindent \textbf{Cross-Domain Mixing}: We investigate the impact of the mixing direction for IMix. Specifically, we compare the effects of source-to-target mixing, in which we cut ground truth instance masks from a source image and paste them to a target image, to our proposed target-to-source mixing, where we rely on the filtered predicted instances from the target domain to augment a source image. As seen in Tab.~\ref{TablePasteDirection} (i), cross-mixing from source-to-target substantially degrades the panoptic performance, specifically the instance segmentation quality. Target-to-target pasting is not included because intra-domain mixing does not apply to UDA and is used in semi-supervised setting.\\
We argue that this performance degradation stems from the incompleteness of the target labels. When training directly on a target image, the pseudo-labels generated from the teacher model are not guaranteed to be exhaustive, i.e. false negative predictions will yield objects on the image with no associated instance label (see the supplement for an example). Within a self-supervised setting, this results in the introduction of confirmation bias that severely degrades performance. On the contrary, we see significant benefits in favor of target-to-source mixing that fully avoids the incompleteness of labels, resulting in $+1.8\%$ mPQ improvement which is mainly attributed to the gains in instance segmentation performance ($+8.2\%$ mAP).
Furthermore in Tab.~\ref{TablePasteDirection} (i) and (ii), we isolate the effects of the mixing task by fixing the mixing direction. Specifically, we compare an inverted ClassMix~\cite{classmix} that cuts and pastes semantic masks from target-to-source, to our proposed IMix strategy that works on an instance level. As seen, the inverted ClassMix slightly underperforms compared to the baseline model and significantly underperforms compared to IMix ($-2.7\%$ mPQ). 
\vspace{-2.5px}

\section{Conclusion}
In this work, we tackle the task of UDA for panoptic segmentation. To this end, we introduce a framework LIDAPS that reduces the domain gap between target and source images by leveraging instance-aware cross-domain mixing. Specifically, we propose a novel mixing strategy IMix, that cuts and pastes confidence-filtered instance predictions from the target to the source domain, and ensures the exhaustiveness of the resulting pseudo-labels while reducing the confirmation bias.
To limit the effects of catastrophic forgetting on the semantic branch, we propose a CLIP-based domain alignment mechanism that employs CLIP embeddings as anchors for both the source and target domain. Combined, our model consistently outperforms existing SOTA on popular UDA panoptic benchmarks. \\
\textbf{Limitations}: We discuss the limitations in the supplement.


%
%
\bibliographystyle{splncs04}
\bibliography{egbib}

\clearpage
\clearpage
\appendix
\section{Overview}
In this section we give an overview on the addressed materials of the appendix. In section~\ref{app:impl} we give implementation details such as the architecture details, hyper-parameter settings, CLIP pre-trained model setting, and the losses we used for training. In section~\ref{app:falseNeg}, we present an example of a false negative pseudo-mask that occurs in source-to-target instance-aware mixing while target-to-source IMix eliminates it. Furthermore, in section~\ref{app:dataset}, we discuss the datasets that we used in our different benchmark reports in the main paper. Moreover, in section~\ref{app:evals}, we explain the different evaluation metrics that we report numbers for in the tables of the paper. In section~\ref{app:qual}, we present additional qualitative results for the LIDAPS model for different instance classes and how they compare to the EDAPS results. In section~\ref{app:quan}, we present a plot bar that illustrates the quantative results for different methods on different benchmarks. In section~\ref{app:abl}, we provide the tables from the Ablation Section of the paper with standard deviation included. Lastly, in section~\ref{app:limit}, we address the limitations of our work.

\section{Implementation Details} \label{app:impl}
\noindent \textbf{Hyper-parameter Settings}: We train our method on a single NVIDIA GeForce RTX 3090. We use an AdamW optimizer with a learning rate of $6 \times 10^{-5}$, a weight decay of 0.01, starting with a linear learning rate warmup for 1.5k iterations, and afterwards a polynomial decay. Furthermore, we train for 50k iterations with a batch size of two, consisting of cropped images of size 512x512. We apply a warmup training phase of 40k iterations and only enable IMix in the last 10k iterations (fine-tuning phase). Generally, we follow the hyper-parameter settings from EDAPS except the IMix confidence threshold and the CLIP loss weight. \\
\noindent \textbf{Architecture}: We use MiT-B5 \cite{segformer} as our encoder backbone (shared by the instance and semantic decoders), MaskRCNN~\cite{maskrcnn} as instance decoder and DAFormer~\cite{daformer} semantic head as the semantic decoder. For CLIP-based domain alignment, we use CLIP~\cite{clip}\footnote{https://huggingface.co/openai/clip-vit-large-patch14} as the pre-trained text encoder. We calculate the CLIP encodings of the categories only once before the start of training. During test time, these components are not needed and thus don't add any computational overhead.\\
\noindent \textbf{IMix Threshold} In Tab.~\ref{TableFilter}, we conduct an hyperparameter study on the confidence-filtering threshold used for cross-domain instance mix-sampling. We evaluate a wide range of thresholds from 0 (no filtering) to 1 (disabling IMix). While we observe a local maxima at 0.75, we note that the method is relatively robust against this selection as it continues to outperform the baseline at a threshold of 0.5 with $+1.9\%$ mPQ and $+6.4\%$ mAP improvements.

\begin{table}[t]
    \centering
    \caption{Hyperparameter study on the confidence-filtering threshold applied to the pseudo-masks for IMix.}
    \vspace{-3.5mm} 
    \scalebox{0.95}{
    \begin{tabular}{l|ccc|cc}    
         Filter & mSQ & mRQ & mPQ &  mIoU& mAP  \\
         \hline
         0 & 73.3$\pm$\tiny{0.1} & 52.1$\pm$\tiny{0.4} & 40.0$\pm$\tiny{0.4} & 59.2$\pm$\tiny{0.8} & 28.7$\pm$\tiny{1.3}\\
         0.25 & 74.0$\pm$\tiny{0.1} & 54.8$\pm$\tiny{0.3}& 42.5$\pm$\tiny{0.3} & 59.3$\pm$\tiny{0.7} & 36.8$\pm$\tiny{1.3}  \\
         0.5  & 74.4$\pm$\tiny{0.5} & 56.5$\pm$\tiny{0.3}& 44.0$\pm$\tiny{0.3} & 59.2$\pm$\tiny{0.7} & 40.8$\pm$\tiny{1.2} \\
         0.75 & \textbf{74.4$\pm$\tiny{0.2}} & \textbf{57.6$\pm$\tiny{0.2}} & \textbf{44.8$\pm$\tiny{0.2}} & \textbf{59.6$\pm$\tiny{0.6}} & \textbf{42.6$\pm$\tiny{0.7}}\\
         1 & 73.9$\pm$\tiny{0.4} & 55.0$\pm$\tiny{0.7} & 42.9$\pm$\tiny{0.6} & 59.6$\pm$\tiny{0.6} & 34.4$\pm$\tiny{0.6}\\         
    \end{tabular}}
    \label{TableFilter}
\vspace{-13px} \end{table}

Thus, we empirically set the IMix confidence threshold at $0.75$ for the settings SYNTHIA $\rightarrow$ Cityscapes and Cityscapes $\rightarrow$ Cityscapes foggy while for SYNTHIA $\rightarrow$ Mapillary and Cityscapes $\rightarrow$ Mapillary we find that the best threshold is 0.9.\\

\noindent \textbf{Losses}
 While the mechanisms we propose (i.e., IMix and CDA) are model agnostic, here we provide detailed mathematical notations of the all losses we used in our end-to-end trainable model, LIDAPS. These formulas have been introduced in prior works~\cite{fasterrcnn, fastrcnn, maskrcnn}, nevertheless, we provide them for the sake of reproducibility and in order to explain the changes that occur to the ground truth supervision of some of these losses when training on IMix augmented images.\\

\begin{equation} 
    \mathcal{L}_\mathrm{pan} = \mathcal{L}_\sem + \mathcal{L}_\mathrm{inst}.
\label{eq:main-inst-sem}
\end{equation}

 As explained in the paper, Eq. \ref{eq:main-inst-sem}, a panoptic loss function consists of two terms; an instance segmentation and a semantic segmentation loss term. Our instance decoder~\cite{maskrcnn} consists of an RPN network and a refinement ($\Refi$) network. Each part has its own losses as shown in Eq.~\ref{eq:instance_dec}. 
  \begin{equation}\label{eq:instance_dec}
     \mathcal{L}_\mathrm{inst}= \mathcal{L}^\mathrm{RPN} + \mathcal{L}^\mathrm{Ref} 
 \end{equation}
 The RPN loss function \cite{fasterrcnn} has two terms, 
 one for the ``objectness'' ($\mathcal{L}_\mathrm{Cls}^\mathrm{RPN}$) 
 and another one for the bounding-box (or region proposal) regression ($\mathcal{L}_\mathrm{Box}^\mathrm{RPN}$)
 loss as seen in Eq.~\ref{eq:rpn-gen}. 
 The RPN takes a predefined set of anchor boxes and the convolution feature map (encoding the input image) as inputs and learns to
 correctly localize objects present in the image.
 For each predicted bounding-box, it predicts an ``objectness'' score 
 indicating whether that box encompasses an object instance or not. 
 The RPN box classification loss $\mathcal{L}_\mathrm{Cls}^\mathrm{RPN}$ is a binary cross-entropy loss which is
 computed between the predicted objectness score $\hat{l}$ and the ground truth objectness label $l$. A class label ``$1$'' denotes that the box region contains an object instance and a label ``$0$'' indicates that there is no object present within the box region.
 This loss encourages the RPN to predict region proposals with high ``objectness'' scores which are later used by the box refinement head for final object detection. 

For the bounding-box regression loss $\mathcal{L}_\mathrm{Box}^\mathrm{RPN}$, an $L1$ loss is used which is computed between the predicted ($\hat{q}$)
and ground truth (${q}$) bounding-box coordinate offsets.
Importantly, the regression loss is only computed for positive predicted boxes \cite{fasterrcnn}.

  \begin{equation} \label{eq:rpn-gen}
     \mathcal{L}^\mathrm{RPN}= \mathcal{L}_\mathrm{Cls}^\mathrm{RPN} + \mathcal{L}_\mathrm{Box}^\mathrm{RPN} 
 \end{equation}

   \begin{equation} \label{eq:rpn-objectness}
     \mathcal{L}_\mathrm{Cls}^{RPN}=  L_{\mathrm{BCE}}\left(\hat{l}, l\right)
 \end{equation}

\begin{equation}\label{eq:rpn-box}
 \mathcal{L}_\mathrm{Box}^\mathrm{RPN} = \lambda_{RPN} \sum_{i\in{x,y,w,r}} L_1(\hat{q}_{i}, q_{i})
\end{equation}

 We use $\textbf{Q}$ to denote the set of ground truth bounding-box offset coordinates when training the student network. As explained in the main paper, LIDAPS is trained on both the source and mixed domain images containing target pseudo-instances. While training on the augmented images (output by IMix), $\textbf{Q}$ represents a union set of the ground truth source bounding-boxes and
confidence-filtered pseudo-bounding-boxes from a target image as shown in Eq. \ref{eq:Q}.
Ground truth bounding-boxes of the source image are denoted by $q^s$, while $q^t$ denotes pseudo-bounding-boxes (predicted by the teacher network) on the target image.
Here, $h_{i}$ is the confidence score predicted for the $i$-th box and the $i$-th mask by the teacher network.

\begin{equation} \label{eq:Q}
\mathbf{Q}=
    \begin{cases}
     \mathbf{Q}^s=\bigcup_{i}q_{i}^s\quad\textrm{if Source}\\
    \mathbf{Q}^s \; \cup \; \bigcup_{i} \mathbbm{1}[h_{i}>\tau]\; q_{i}^t  \quad\textrm{if IMix}
\end{cases}
\end{equation}

The refinement network consists of a box-head and a mask-head following FastRCNN~\cite{fastrcnn}. As seen in Eq.~\ref{eq:refine-gen}, the box-head is trained using a box classification loss $\mathcal{L}_\mathrm{Cls}^\mathrm{Ref}$
and a box regression loss $\mathcal{L}_\mathrm{Box}^\mathrm{Ref}$, 
while the mask-head has a mask segmentation loss $\mathcal{L}_\mathrm{Mask}^\mathrm{Ref}$. 

  \begin{equation} \label{eq:refine-gen}
     \mathcal{L}^\mathrm{Ref}=  \mathcal{L}_\mathrm{Cls}^\mathrm{Ref} + \mathcal{L}_\mathrm{Box}^\mathrm{Ref} + \mathcal{L}_\mathrm{Mask}^\mathrm{Ref}  
 \end{equation}

The box-head takes as inputs the RoIAlign~\cite{maskrcnn} features and the region proposals output by the RPN network, 
and predicts refined bounding-boxes and their classification scores.
The classification scores are the softmax probability scores
for all the thing classes plus a background class($C_{\mathrm{things}}$+1). 

Similar to the RPN, the box-head has a box classification loss 
$\mathcal{L}_\mathrm{Cls}^\mathrm{Ref}$
and a box regression loss
$\mathcal{L}_\mathrm{Box}^\mathrm{Ref}$.
The box classification loss is computed between
predicted per-class probabilities $P_{cl}$  and the ground truth class label $u \in \mathbf{U}$ for the predicted box
as shown in Eq. \ref{eq:refine-class}.
Unlike RPN, where the box classification loss is a binary cross-entropy loss,
$\mathcal{L}_\mathrm{Cls}^\mathrm{Ref}$ is a categorical cross-entropy loss
for multi-class classification.

 \begin{equation}\label{eq:refine-class}
\mathcal{L}_\mathrm{Cls}^\mathrm{Ref} = L_{CE}(P_{cl}, u)
\end{equation}

The box regression loss is computed between the 
predicted bounding-box $\hat{v}_{u,i}$ and the ground truth bounding-box $v_{i}$
as shown in Eq. \ref{eq:refine-reg}. The predicted bounding-box $\hat{v}_{c,i}$ by the box-head is for the class $c \in C_{things}$. Having predictions for all classes mitigates the competition between the classes. 

 \begin{equation}\label{eq:refine-reg}
\mathcal{L}_\mathrm{Box}^\mathrm{Ref} = \lambda_{Ref}\sum_{i\in{x,y,w,r}}L1(\hat{v}_{u,i},v_{i})
\end{equation}

Similar to RPN training, 
the box-head is trained on both source and target domain bounding-boxes \textbf{Q}.
While training on source images, we use the ground truth source bounding-boxes,
and for training on augmented images (output by IMix),
we use a union set of the ground truth source and pseudo bounding-boxes as in Eq. \ref{eq:Q}.
 
$\mathbf{U}$ denotes the ground-truth bounding-box class labels. When training the student network on the source domain images, we use the source ground-truth labels $\mathbf{U}^s$ and while training on the augmented images generated by IMix, $\mathbf{U}$ represents a union set of ground truth source bounding boxes
and confidence-filtered pseudo bounding-box class labels as shown in 
Eq. \ref{eq:U}.

\begin{equation} \label{eq:U}
\mathbf{U}=
    \begin{cases}
     \mathbf{U}^s=\bigcup_{i}u_{i}^s\quad\textrm{if Source}\\
    \mathbf{U}^s \; \cup \; \bigcup_{i} \mathbbm{1}[h_{i}>\tau]\; u_{i}^t  \quad\textrm{if IMix}
\end{cases}
\end{equation}

 For the RPN and box refinement head losses, we set the loss weights $\lambda_{\RPN}$ and $\lambda_{\Refi}$ to 1.0. 

The mask-head predicts $C_{things}$ masks of dimension $w \times h$ for each of the RoIs. Each predicted mask, $\hat{m}_{c}$, is for an RoI and a specific class. This mitigates the competition in between the classes. Each predicted mask is associated to a ground truth mask $m \in \mathbf{Masks}$ according to maximum IoU. When training with IMix, $\mathbf{Masks}$ contains confidence-filtered pseudo-masks $ m^t $ from the target as well as ground truth masks from the source $m^{s}$ as shown in Eq.~\ref{eq:Masks}. \\

\begin{equation} \label{eq:Masks}
\mathbf{Masks}=
    \begin{cases}
     \mathbf{Masks}^s=\bigcup_{i}m^s_{i}\quad\textrm{if Source}\\
    \mathbf{Masks}^s \; \cup \; \bigcup_{i} \mathbbm{1}[h_{i}>\tau]\; m_{i}^t  \quad\textrm{if IMix}
\end{cases}
\end{equation}

Eq.~\ref{eq:mask} indicates the binary cross-entropy loss computed between the predicted $\hat{m}$ 
and ground truth masks $m$,
where $u \in C_{things}$ denotes the ground truth class label for the predicted mask.
 \begin{equation}\label{eq:mask}
\frac{1}{w \times h}\sum_{1\le i,j\le h} m_{i,j} \log (\hat{m}_{u,i,j})  + (1-m_{i,j}) \log(1-\hat{m}_{u,i,j}) \text{.}
 \end{equation} 

Before training with IMix, we first pass the target images through the instance decoder of the teacher network $\theta_{\mathrm{inst}}$ in order to gather the predictions which serve as pseudo-class labels, pseudo-masks, pseudo-bounding-boxes for the student network training.
 The instance decoder of the teacher network provides per-class probabilities for each of the regions of interest. We use the class with the highest probability as the pseudo-label for the i-th ROI which is shown below:
 \begin{equation}
   y^t_{\mathrm{inst}_{i}} = \left[\argmax_{c'} (\theta_\mathrm{inst}(x^{(t)}))_{i}\right]
\end{equation}

The semantic loss on the source domain is explained in Eq.~\ref{eq:sem-source} which defines a categorical cross-entropy loss on the predicted class probability for each pixel.

 \begin{equation} \label{eq:sem-source}
     \mathcal{L}^{s}_\textrm{sem}(\hat{y}_\textrm{sem}^{s}, y_{\textrm{sem}}^{s}) = -\sum_{i,j,c} \left(y_\textrm{sem}^{s} \log(\hat{y}_\textrm{sem}^{s})\right)_{i,j,c}
 \end{equation}

Following~\cite{edaps}, the self-supervised semantic loss applied to the semantic-aware mixed image~\cite{classmix} is shown in Eq.~\ref{eq:sem-mixed}. 
The augmented or mixed image generated using the ClassMix~\cite{classmix} contains pixels from both the source and the target domain images.
For the source pixels, we compute the categorical cross-entropy
loss between the predicted and ground truth semantic class labels.
For the target pixels, we compute a weighted categorical cross-entropy
loss as it takes into account the confidence of the pseudo-semantic class labels predicted by the teacher network.

Thus, $k^t_{(i,j)}$ defines the per-pixel confidence score for every pseudo-label predicted by the teacher network.  $y^t_{\sem}$ is the per-pixel pseudo-label as shown in Eq.~\ref{sem-pseudo} where $\theta_{\sem}$(the semantic decoder of the teacher) predicts per-pixel-class probabilities.

\begin{equation} \label{eq:sem-mixed}
  \mathcal{L}^{ss}_{sem}(\hat{\tilde{y}}_{\textrm{sem}}, \tilde{y}_{\textrm{sem}}) = 
  \begin{cases}
    \mathcal{L}^s_{\sem}(\hat{\tilde{y}}_\sem, y^s_\sem),  \\  \hphantom{ozan} \textrm{ if } \mathbf{M}_{\textrm{sem}}^{(i,j,c)} = 1,
    \\
    -\sum
    k^t_{(i,j)}
    \left( 
      y^t_{\sem} \log(\hat{\tilde{y}}_{\sem})
    \right)_{(i,j,c)}, \\  \hphantom{ozan} \textrm{otherwise } 
    \end{cases}
\end{equation}

\begin{equation}\label{sem-pseudo}
   y^t_{\sem} = \left[\argmax_{c'} (\theta_\textrm{sem}(x^{(t)}))_{i,j}\right]
\end{equation}

\noindent \textbf{EDAPS*}:  This baseline follows the same setting as EDAPS~\cite{edaps} except that it does not include the features distance regularizor (FD) that EDAPS applies during training. FD uses ImageNet features as an anchor in order to hinder the learned encoder from forgetting the knowledge it starts out with when initialized with a pre-trained ImageNet encoder. The regularizor is explained in Eq.~\ref{eq:fd}. Noteworthy is that FD is applied only on source images in areas corresponding to thing classes. In Table \ref{TableFD} we show how the inclusion of FD hinders the performance of our method and thus explains why this component was removed from our experiments. We speculate that this is because the embedding spaces of ImageNet and CLIP are not aligned, therefore, aligning with both gives rise to a drop in performance. Additionally, EDAPS* is trained for 50k iterations instead of 40k which is the duration of training reported for EDAPS. In Table~\ref{Table50kEDAPS}, we compare EDAPS with LIDPAS, both trained for 50k iterations. We can see that LIDAPS persists on beating EDAPS on three different benchmarks.

\begin{equation}\label{eq:fd}
\mathcal{L}_\mathrm{FD}=\lVert \mathrm{Enc}_\mathrm{ImgNet}(x^s) - \mathrm{Enc}_\theta(x^s)  \rVert
\end{equation}

\begin{table}[h]
    \centering
     \caption{ Ablation study on EDAPS and LIDAPS in an equalized setting where EDAPS is trained for 50k iterations on three different benchmarks.}
     \vspace{-2.5mm} 
     \scalebox{0.9}{
    \begin{tabular}{c|ccc|cc}
          Method & mSQ & mRQ & mPQ & mIoU & mAP \\
         \hline
         \multicolumn{5}{c}{SYNTHIA $\rightarrow$ Cityscapes}\\
         \hline
         EDAPS &  72.4$\pm$\tiny{0.4} &53.2$\pm$\tiny{1.0}  & 40.8$\pm$\tiny{0.9} &57.5$\pm$\tiny{0.7}  & 33.7$\pm$\tiny{0.6} \\
        LIDAPS & \textbf{74.4$\pm$\tiny{0.28}} & \textbf{57.6$\pm$\tiny{0.294}} & \textbf{44.8$\pm$\tiny{0.2}} & \textbf{59.6$\pm$\tiny{0.6}} & \textbf{42.6$\pm$\tiny{0.7}} \\
        \hline

         \multicolumn{5}{c}{SYNTHIA $\rightarrow$ Mapillary Vistas}\\
        \hline
         EDAPS &  72.9$\pm$\tiny{0.4} &46.1$\pm$\tiny{0.2}  & 36.6$\pm$\tiny{0.2} &55.4$\pm$\tiny{4.1}  & 32.8$\pm$\tiny{0.3} \\
        LIDAPS & \textbf{73.9$\pm$\tiny{1.9}} & \textbf{47.7$\pm$\tiny{0.2}} & \textbf{38.0$\pm$\tiny{0.2}} & \textbf{58.8$\pm$\tiny{0.5}} & \textbf{38.7$\pm$\tiny{0.2}} \\
        \hline

        \multicolumn{5}{c}{Cityscapes $\rightarrow$ Cityscapes foggy}\\
        \hline 
         EDAPS &  79.2$\pm$\tiny{0.1} &71.2$\pm$\tiny{0.0}  & 57.3$\pm$\tiny{0.2} &83.0$\pm$\tiny{0.6}  & 60.4$\pm$\tiny{0.4} \\
        LIDAPS & \textbf{80.2$\pm$\tiny{0.1}} & \textbf{73.2$\pm$\tiny{0.6}} & \textbf{59.6$\pm$\tiny{0.6}} & \textbf{87.1$\pm$\tiny{0.7}} & \textbf{65.3$\pm$\tiny{0.6}} \\

        
    \end{tabular}}
   
    \label{Table50kEDAPS}
\end{table}

\begin{table}[h]
    \centering
     \caption{ Ablation study on the FD component. We include feature distance (FD) in our proposed LIDPAS model (LIDAPS$_{\mathcal{FD}}$) and compare its performance to LIDAPS.}
     \vspace{-2.5mm} 
     \scalebox{0.8}{
    \begin{tabular}{c|ccc|cc}
          Method & mSQ & mRQ & mPQ & mIoU & mAP \\
         \hline
         LIDAPS$_{\mathcal{FD}}$ &  74.0$\pm$\tiny{0.3} & 56.1$\pm$\tiny{1.3} & 43.7$\pm$\tiny{0.9} & 58.6$\pm$\tiny{0.8} & 40.3$\pm$\tiny{0.9} \\
        LIDAPS & \textbf{74.4$\pm$\tiny{0.28}} & \textbf{57.6$\pm$\tiny{0.294}} & \textbf{44.8$\pm$\tiny{0.2}} & \textbf{59.6$\pm$\tiny{0.6}} & \textbf{42.6$\pm$\tiny{0.7}} \\
    \end{tabular}}
   
    \label{TableFD}
\vspace{-10px}\end{table}

\section{False Negatives} \label{app:falseNeg}
As explained in the paper, when instance-aware mixing is done from source to target, exhaustive pseudo-masks for the target instances are not guaranteed. In Fig.~\ref{fig:falseNegatives}, we show an example where in (c) confidence-filtered target instances are pasted onto the source image while in (d) all ground truth source instances are pasted on to the target image. In Fig.~\ref{fig:falseNegatives}(c), we can see that the target instances all have a pseudo-mask while in Fig.~\ref{fig:falseNegatives}(d), the encircled instance (the truck) in red does not have a corresponding pseudo-mask which is indicative of a false negative. When going from target to source, only the instances with a corresponding pseudo-mask are copy and pasted. Thus, inherently, all of the pasted target instances have a corresponding pseudo-mask. On the other hand, when remaining in the target image, target instances with absent pseudo-masks remain.

\begin{figure*}[t]
    \centering
    \includegraphics[width=\textwidth]{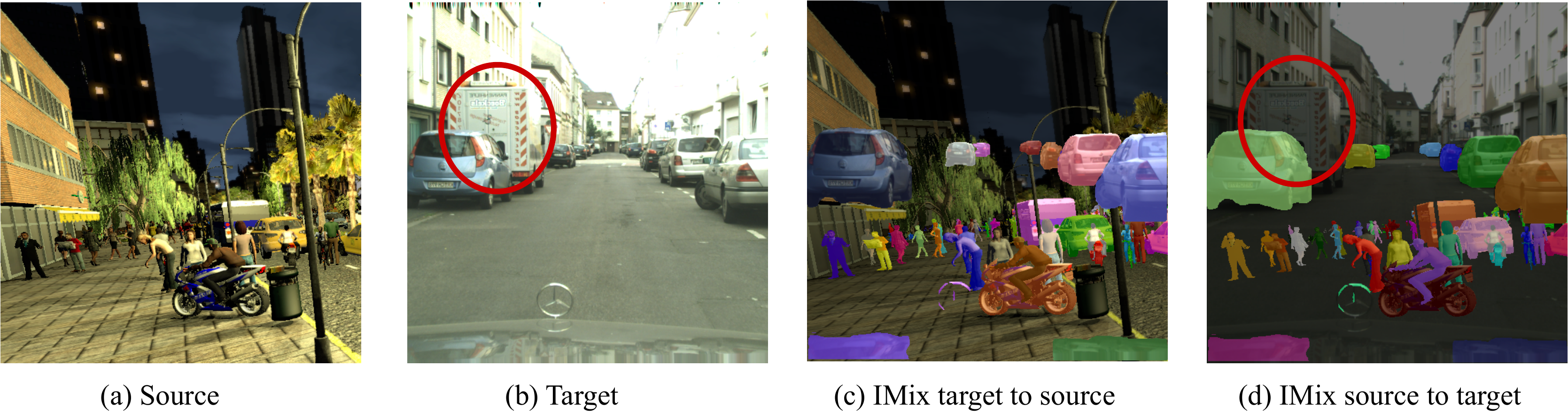}
    \caption{When using IMix to paste source instances from source to target (c), exhaustive pseudo-masks for the target instances is not guaranteed. For instance, in (d) the truck has no pseudo-mask. In (c), this exhaustiveness is guaranteed because only target instances with predicted pseudo-masks are pasted onto the source image. Thus, training on samples mixed from target to source allows the model to learn on supervised sets with no false negative examples.}
    \label{fig:falseNegatives}
\vspace{-10px}\end{figure*}

\section{Datasets} \label{app:dataset}
We evaluate our method on the popular panoptic UDA benchmarks. For synthetic-to-real adaptation, we use SYNTHIA~\cite{synthia} as the source domain which contains 9,400 synthetic images. For the target domain, we use the Mapillary Vistas~\cite{mapillary} dataset and Cityscapes~\cite{cityscapes}. Cityscapes contains 2,975 training images and 500 validation images, while Mapillary Vistas contains 18,000 training images and 2,000 validation images. For real-to-real adaptation, we use two different benchmarks. First, we train with Cityscapes as the source and Mapillary Vistas as the target domain, and second, we train with Cityscapes as the source and the adverse weather dataset Foggy Cityscapes~\cite{foggy} as the target domain.

\section{Evaluation Metrics} \label{app:evals}
We report the mean panoptic quality (mPQ) for panoptic segmentation, which measures both the semantic quality (SQ) and the recognition quality (RQ). To highlight the individual task performances, we further report the mIoU for semantic segmentation over 20 classes, and mAP for instance segmentation over 6 \emph{thing} classes. All reported values are the averaged scores over three runs with three different seeds (1, 2, 3).

\section{Additional Qualitative Results} \label{app:qual}
In this section, we provide additional qualitative panoptic segmentation results in Fig. \ref{fig:add-qty-results}.
\begin{figure*}[h]
    \centering
    \includegraphics[width=\textwidth]{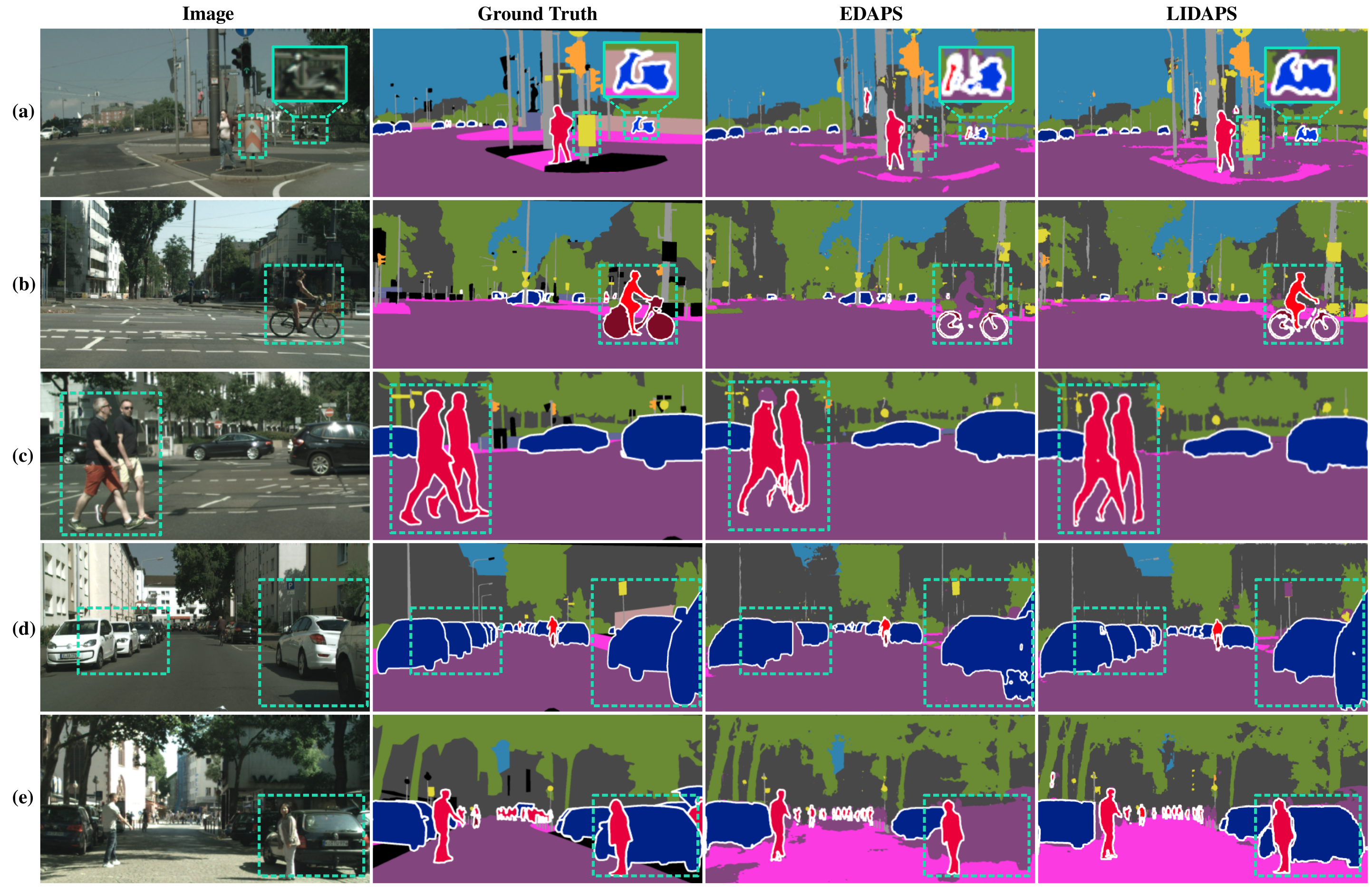}
    \caption{Additional qualitative results on SYNTHIA $\rightarrow$ Cityscape UDA benchmark comparing EDAPS~\cite{edaps} to our proposed LIDAPS. Our proposed LIDAPS model predicts improved semantic and instance segmentation for several classes including ``motor-bike'' (a), ``rider'' (b), ``person'' (c) and ``car'' (d,e). }
    \label{fig:add-qty-results}
\vspace{-10px}
\end{figure*}

\section{Additional Qualitative Results} \label{app:quan}
In Fig. \ref{fig:intro-bar-plot}, we display the quantative results of LIDAPS and other UDA panoptic methods for different benchmarks on a bar plot.
   \begin{figure*}[h]
     \centering
     \includegraphics[width=\textwidth]{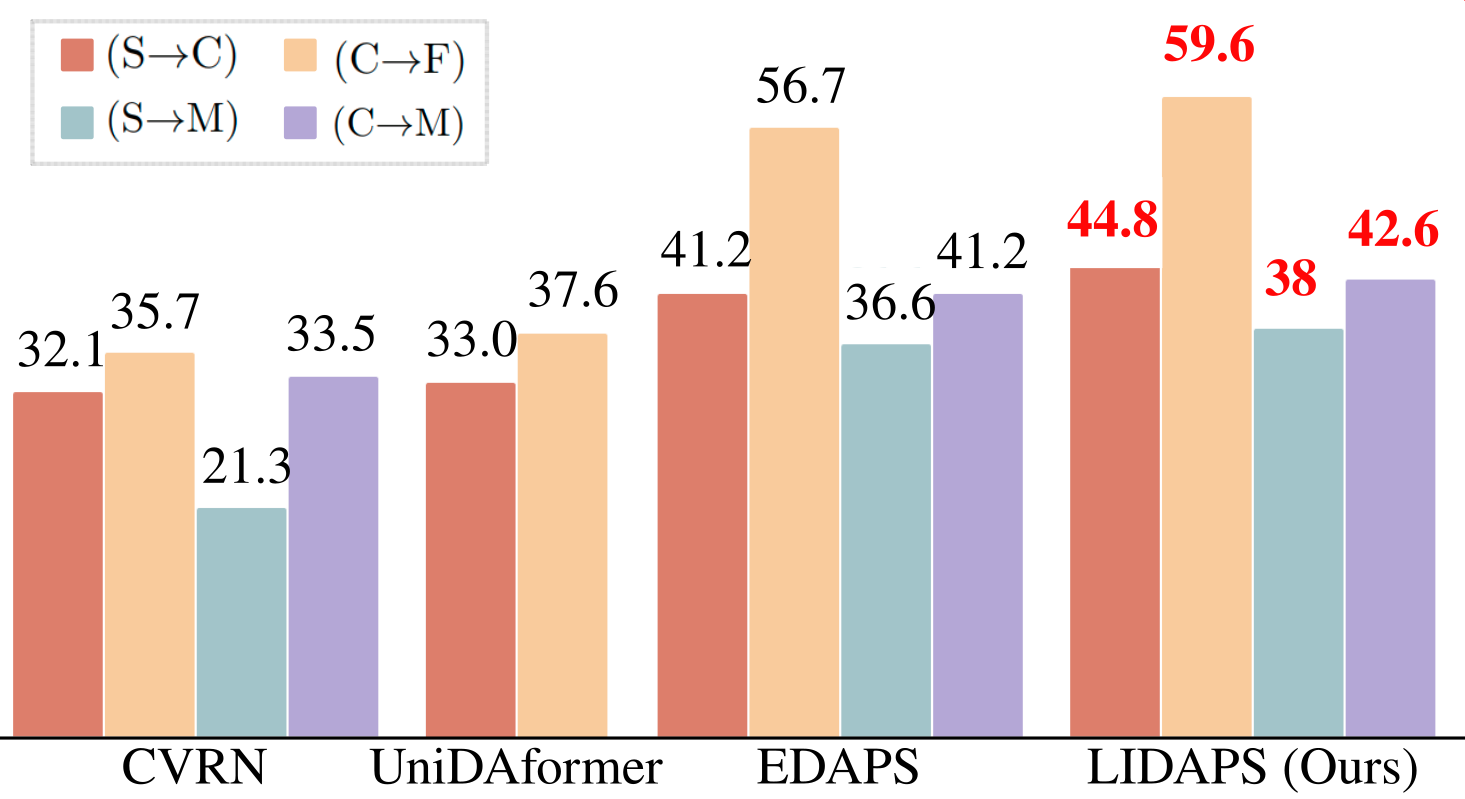}
     \vspace{-6.5mm} 
     \captionof{figure}{The two main contributions, IMix and CDA help in improving the UDA panoptic (mPQ) over the SOTA on four UDA panoptic segmentation benchmarks
     S$\rightarrow$C: SYNTHIA to Cityscapes, C$\rightarrow$F: Cityscapes to Foggy Cityscapes, S$\rightarrow$M: SYNTHIA to Mapillary and C$\rightarrow$M: Cityscapes to Mapillary. }
     \label{fig:intro-bar-plot}
  \end{figure*}

\section{Ablation studies including Standard Deviations} \label{app:abl}
In this section, we provide Table~\ref{TableAblateAll} and Table~\ref{TablePasteDirection} for the ablation studies of the main paper where we additionally include the standard deviation of the results each conducted for three rounds.
\begin{table*}[t]
    \centering
     \caption{ Ablation study on proposed modules. Starting from a baseline EDAPS*, we individually introduce our instance-aware cross-domain mixing (IMix) and CLIP-based domain alignment (CDA). Each experiment is run three times.}
     \vspace{-2.5mm} 
     \scalebox{0.9}{
    \begin{tabular}{ccc|ccc|cc}
         EDAPS$^*$ & IMix & CDA & mSQ & mRQ & mPQ & mIoU & mAP \\
         \hline
         \ding{51}  &  &   &  72.3$\pm$\tiny{0.2} &53.3$\pm$\tiny{0.8}  & 41.0$\pm$\tiny{0.4} &58.0$\pm$\tiny{0.2}  & 34.1$\pm$\tiny{1.0} \\
          \ding{51}  & \ding{51} & & 73.0$\pm$\tiny{0.0} & 54.7$\pm$\tiny{0.8} & 42.3$\pm$\tiny{0.6}  &57.7$\pm$\tiny{0.3} & 39.5$\pm$\tiny{2.3}  \\
         \ding{51}  &  &\ding{51} &  73.9$\pm$\tiny{0.3} & 55.0$\pm$\tiny{0.6}  & 42.9$\pm$\tiny{0.6} & 59.6$\pm$\tiny{0.6} & 34.4$\pm$\tiny{0.6}  \\
         \ding{51} &  \ding{51}  & \ding{51} & \textbf{74.4$\pm$\tiny{0.2}} & \textbf{57.6$\pm$\tiny{0.2}} & \textbf{44.8$\pm$\tiny{0.2}} & \textbf{59.6$\pm$\tiny{0.6}} & \textbf{42.6$\pm$\tiny{0.7}}\\
         
    \end{tabular}}
   
    \label{TableAblateAll}
\vspace{-12px}\end{table*}

\begin{table*}[t]
    \centering
     \caption{Ablation study on mixing strategy for panoptic segmentation comparing (i) the mixing direction when applying IMix, (ii) the effects of ClassMix when applied from target-to-source as opposed to source-to-target. The baseline is EDAPS*+CDA. Each experiment is run three times. S stands for source while T stands for target.}
     \vspace{-2.5mm} 
     \scalebox{0.9}{
    \begin{tabular}{ll|cc|ccc|cc}
        & Method & Copy & Paste & mSQ & mRQ & mPQ & mIoU & mAP \\
        \hline
        & Baseline & - & - & 73.9$\pm$\tiny{0.3} & 55.0$\pm$\tiny{0.6}  & 42.9$\pm$\tiny{0.6} & 59.6$\pm$\tiny{0.6} & 34.4$\pm$\tiny{0.6}\\
         \hline
         \multirow{2}{*}{(i)}  & \multirow{2}{*}{+ IMix} & S & T & 62.0$\pm$\tiny{3.3} & 37.6$\pm$\tiny{0.6} & 29.3$\pm$\tiny{0.5} & 56.2$\pm$\tiny{0.7} & 1.9$\pm$\tiny{1.9}\\
          & & T & S & \textbf{74.4$\pm$\tiny{0.2}} & \textbf{57.6$\pm$\tiny{0.2}} & \textbf{44.8$\pm$\tiny{0.2}} & \textbf{59.6$\pm$\tiny{0.6}} & \textbf{42.6$\pm$\tiny{1.7}}\\
        \hline
        (ii) & + ClassMix~\cite{classmix} & T & S & 73.5$\pm$\tiny{0.2} & 53.9$\pm$\tiny{0.7} & 42.1$\pm$\tiny{0.6} & 58.6$\pm$\tiny{0.8} & 34.8$\pm$\tiny{0.9}\\
    \end{tabular}  } 
    \label{TablePasteDirection}
\vspace{-6px} \end{table*}

\section{Limitations} \label{app:limit}
Depending on the source and target domain, the threshold for pseudo-mask confidence filtering needs to be manually found with experiments. Thus, we show that this threshold is different on different benchmarks. In future work, we will explore the prediction of the threshold using a jointly trained neural network.
Furthermore, during the refinement phase where IMix is enabled (last 10k iterations), we are adding one forward pass and one backward pass to each iteration which increases the runtime.

\end{document}